# A Few Good Counterfactuals: Generating Interpretable, Plausible & Diverse Counterfactual Explanations

**Barry Smyth**[*] **& Mark T. Keane**

Insight SFI Centre for Data Analytics,
VistaMilk SFI Research Centre,
School of Computer Science,
University College Dublin, Dublin, Ireland
{barry.smyth, mark.keane}@ucd.ie

## Abstract

Counterfactual explanations provide a potentially significant solution to the Explainable AI (XAI) problem, but good, "native" counterfactuals have been shown to rarely occur in most datasets. Hence, the most popular methods generate synthetic counterfactuals using "blind" perturbation. However, such methods have several shortcomings: the resulting counterfactuals (i) may not be valid data-points (they often use features that do not naturally occur), (ii) may lack the sparsity of good counterfactuals (if they modify too many features), and (iii) may lack diversity (if the generated counterfactuals are minimal variants of one another). We describe a method designed to overcome these problems, one that adapts "native" counterfactuals in the original dataset, to generate sparse, diverse synthetic counterfactuals from naturally occurring features. A series of experiments are reported that systematically explore parametric variations of this novel method on common datasets to establish the conditions for optimal performance.

## 1 Introduction

Imagine your research paper has been reviewed by an AI model that provides post-hoc explanations for its decisions. The system might explain a rejection using a counterfactual statement: *"if the paper was written more clearly and the evaluation used more datasets, then it would have been accepted"*. This proposes an alternative outcome (acceptance) if two aspects of the paper had been different (clearer writing *and* a stronger evaluation). This explanation is just one of many possible alternative counterfactuals for explaining a rejection. For example, another counterfactual might emphasise different features of the paper (related work and technical detail), with the statement: *"if the paper had included a better treatment of related work and provided a clearer technical account of the main algorithm, then it would have been accepted"*. While we hope that neither of these criticisms will be applied to the present work, they show how useful counterfactuals can be in providing a causal focus for how an alternative outcome *could* be achieved [Miller, 2019; Byrne, 2019], rather than merely explaining why a particular outcome *was* achieved [Doyle *et al.*, 2004; Kenny and Keane, 2019]. Accordingly, in recent years, there has been an explosion of research on how counterfactuals can be used in Explainable AI (XAI) [Adadi and Berrada, 2018; Verma *et al.*, 2020] and algorithmic recourse [Karimi *et al.*, 2020].

From a machine learning perspective, counterfactuals correspond to the *nearest unlike neighbours* (NUNs) of a target problem to be explained [Dasarathy, 1994; McKenna and Smyth, 2000; Doyle *et al.*, 2004]. NUNs can be cast as *native counterfactuals*, but not every NUN will make a good counterfactual: a NUN may not be similar enough to the target problem to be a useful explanation, or it may involve too many feature differences to be easily understood (i.e., lacking sparsity [Wachter *et al.*, 2017]); it has argued, for psychological reasons, that good counterfacutals should have no more than 2 feature differences [Keane and Smyth, 2020]. Also, it might not be possible to find a NUN that emphasises the right sort of differences, those that are actionable in practice. Indeed, [Keane and Smyth, 2020] have shown that good, native counterfactuals are rare (95% of datasets examined had <1% of natives with ≤2 differences). Hence, most counterfacutal solutions generate *synthetic* counterfactuals, rather than trying to find them in the dataset. As such we distinguish between the generation of two distinct classes of synthetic counterfactuals:

- *Endogenous counterfactuals* are generated from naturally occurring feature values; that is, they use feature values that exist in other (native) instances

- *Exogenous counterfactuals* are generated in ways

---

[*]Contact Author

that are not guaranteed to use naturally occurring feature values; e.g., they may rely on interpolations of existing feature values.

This distinction is important because it has implications for the plausiblity and actionabllility of generated counterfactuals; e.g., an explanation saying that a property needs 2.32 bedrooms for a higher sale-price is neither plausible nor actionable. Though most current counterfactual generation techniques produce exogenous counterfactuals, here we focus on endogenous counterfactual methods by adapting "native" counterfactuals (NUNs in the original dataset) to generate plausible contrastive explanations. Recently, [Keane and Smyth, 2020] (henceforth, KS20) reported an endogenous nearest-neighbour algorithm that generates counterfactuals from the existing features of a target problem and a suitable nearby NUN (when one exists). The present paper generalises this technique and shows that this new method delivers significant improvements in counterfactual quality, across many datasets on key evaluation metrics.

## 2 Related Work

As AI systems are becoming more widespread in our everyday lives, the need for fairness [Russell *et al.*, 2017], transparency [Larsson and Heintz, 2020], explainability is becoming more important [Adadi and Berrada, 2018; Guidotti *et al.*, 2018; Kusner and Loftus, 2020; Muhammad *et al.*, 2016]; indeed, some governmental regulations (such as GDPR) now call for mandatory explanations for AI-based decisions [Goodman and Flaxman, 2017]. At the same time, those machine learning approaches that have proven to be so effective in real-world tasks (e.g. deep neural networks), appear to be among the most difficult to explain [Gunning, 2017]. One approach to this problem is to cast such black-box models as white-box ones and then to use the latter to explain the former; for example, using post-hoc feature-based (e.g., as in LIME [Ribeiro *et al.*, 2016]) or example-based explanations (e.g., as in twin-systems [Kenny and Keane, 2019; Gilpin *et al.*, 2018]). Counterfactual explanations have emerged as another popular post-hoc option as they (are argued to) offer psychological, technical and legal benefits [Miller, 2019; Byrne, 2019; Wachter *et al.*, 2017; Mittelstadt *et al.*, 2019]. In this literature, there is a consensus that *good counterfactuals* should be:

- *Available*: for a majority of target problems encountered, giving a high degree of explanation coverage.
- *Similar*: maximally similar to the target problem, to be understandable to end-users.
- *Sparse*: differ in as few features as possible from the target problem, to be easily interpreted.
- *Plausible*: contain feature values or combinations of feature values that make sense (e.g., preferably from known instances).
- *Diverse*: using different features, to provide explanations from different perspectives (when multiple alternatives are required)

Native counterfactuals – such as existing NUNs – are plausible by definition, but natives that are similar and sparse are rare [Keane and Smyth, 2020]. Hence, most counterfactual methods generate synthetic, exogenous counterfactuals. The seminal work of [Wachter *et al.*, 2017] proposed an optimisation approach, that generates a new counterfactual $p'$ for a target problem $p$ by perturbing the features of $p$, until a class change occurs, in a manner that minimises $d(p,p')$, the distance between $p$ and $p'$. While this approach can generate a $p'$ that is very similar to $p$, these "blind" perturbation methods can still generate counterfactuals that lack sparsity [McGrath *et al.*, 2018], offer limited diversity [Mothilal *et al.*, 2020] and can involve invalid, out-of-distribution data-points [Laugel *et al.*, 2019]. Hence, [Dandl *et al.*, 2020] have proposed modifications to the loss function to minimise the number of different features between $p$ and $p'$ (*diffs(p,p')*). And, [Mothilal *et al.*, 2020] have extended the optimisation function to deal with diversity; so that for given $p$, the set of counterfactuals produced minimises the distance and feature differences within the set, while maximising the range of features changed across the set. However, to deal with the out-of-distribution problem, others have explored fundamentally-different endogenous approaches, in which known instances are exploited more directly to generate good counterfactuals [Keane and Smyth, 2020; Laugel *et al.*, 2019; Poyiadzi *et al.*, 2020].

Notably, [Keane and Smyth, 2020] (KS20) proposed an endogenous technique for generating counterfactuals by adapting native counterfactuals in the dataset. They defined a *good* counterfactual to be one with ≤2 feature differences with respect to a target problem, $p$. If a good native counterfactual did not exist for $p$ then KS20 *reused* the feature differences from a *good native counterfactual* that existed for some other problem $q$ similar to $p$. The differences between $q$ and its good, native-counterfactual ($q'$) serve as a *template* for generating the new (good) counterfactual for $p$, using only existing, naturally-occurring feature-values. Even though good native counterfactuals are rare, because they can be adapted in many different ways, KS20 showed their technique could generate counterfactuals that were very similar to target problems, while remaining sparse (i.e., ≤2 feature differences) and plausible. However, although KS20 demonstrated the feasibility of an endogenous approach (on a small set of parametric variations), it was limited to generating singleton counterfactuals thereby limiting coverage, similarity, and diversity.

Using KS20 aqs a starting point, the present work describes a novel endogenous method that significantly extends this previous work. First, it proposes a more elegant and general *k*-NN approach, capable of generating superior counterfactual candidates from the *k* nearest native-counterfactuals, rather than just a single nearest-counterfactual. Second, this new method is capable of generating diverse sets of good counterfactuals. Third, in systematic tests varying key parameters, it is shown to consistently and significantly improve coun-terfactual coverage and similarity, when compared to a KS20-baseline, for a respectable number of datasets.

## 3 Generating Good Counterfactuals

The work of [Keane and Smyth, 2020] (KS20) assumed an explanation context in which an underlying decision model ($M$; e.g. a deep learner) was making predictions to be explained by a twinned counterfactual generator using $k$-NN (see [Kenny and Keane, 2019]). Their method aims to generate *good counterfactuals* ($\leq 2$ feature differences) to explain the prediction of a target problem, $p$. Given a set of training cases/instances, $I$, the approach relies on the reuse of an existing good (native) counterfactual, represented as a so-called *explanation case* (XC). An explanation case, $xc$, contains a target problem instance, $x$, and a nearby NUN, $x'$, with no more than $d = 2$ feature differences; see Equations 1-5:

$$NUN(x, x') \iff class(x) \neq class(x') \quad (1)$$

$$matches(x, x') = \{f \; \epsilon \; x \mid x.f \approx x'.f\} \quad (2)$$

$$diffs(x, x') = \{f \; \epsilon \; x \mid x.f \not\approx x'.f\} \quad (3)$$

$$xc_d(x, x') \iff NUN(x, x') \land \mid diffs(x, x') \mid \leq d \quad (4)$$

$$XC_d = \{xc_d(x, x') \; \forall \; x, x' \epsilon \; I\} \quad (5)$$

Each explanation case, $xc_d$, is associated with a set of *match* features, whose values are (approximately) equal in $x$ and $x'$, and a set of *difference* features, whose values are different. Thus, an $xc$[1] acts as *template* for generating new counterfactuals, by identifying features that can be adapted (the difference features) and those which should remain invariant (the match features).

To generate a good counterfactual for some target problem, $p$, the KS20 method first identifies the nearest neighbour, $xc$, based the similarity between $p$ and the target problem of $xc$, namely $xc.x$, and then constructs a new counterfactual, $cf$ from the feature values of $p$ and $xc.x'$ (the good counterfactual for $x$). The values of match features in $p$ are transferred to $cf$ along with the values of the difference features from $x'$; see lines 14-18 in Algorithm 1. If the predicted class of $cf$ differs from the class of $p$ ($class(cf) \neq class(p)$), then $cf$ is considered to be a valid (good) counterfactual for $p$. If $class(cf) = class(p)$, then sn alternative counterfactual can be generated using the values of $xc$'s difference features from the nearest-like-neighbours of $x'$.

### 3.1 Reusing Multiple Explanation Cases

In essence, KS20 proposed a *1NN* approach to counterfactual generation, as a *single* nearest explanation-case was used as a template for the counterfactual. It is natural to consider a $k$-NN extension, where $k$ nearest-neighbour explanation-cases are reused, each providing a different set of counterfactual candidates (see Algorithm 1). Briefly, given a target problem, $p$, a set of $k$ nearest explanation-cases are selected (lines 5-7), based on the similarity between $p$ and their target problems ($xc.x$); using a euclidean distance metric with feature values scaled to unit variance. For each nearest $xc$, we generate a set of candidate counterfactuals based on the features of it's NUN or good counterfactual ($xc.x'$) and the like neighbours of $xc.x'$ (lines 8-13). A counterfactual is generated for these NUNs (lines 14-18) and validated by comparing the predicted class to the NUN class (lines 18-20); further discussion of this validation step, which differs from KS20, see Section 3.2. Thus the time complexity of generating a set of counterfactuals for a given $p$ is $O(km)$ where $m$ is the average number of like-neighbours of the NUNs in explanation cases.

---

[1]For convenience, we drop the $d$ without loss of generality.

---

**Given** : $p$, target problem;
  $I$, training instances;
  $XC_d$, explanation cases for $d$;
  $k$, number of XCs to be reused;
  $M$, underlying (classification) model.
**Output:** $cfs$, valid counterfactuals for $p$.

1 **def** gen-kNN-CFs($p, I, XC_d, k, M$):
2     $xcs \leftarrow getXCs(p, XC_d, k)$
3     $cfs \leftarrow \{genCFs(p, xc, I, M) \mid xc \; \epsilon \; xcs\}$
4     **return** $cfs$

5 **def** getXCs($p, XC, k$):
6     $XC' \leftarrow \{xc \in XC \mid class(xc.x) = class(p)\}$
7     $xcs \leftarrow sort(XC', key = sim(xc.x, p))$
8     **return** $xcs[:k]$

9 **def** genCFs($p, xc, I, M$):
10     $nun \leftarrow \{xc.x'\}$
11     $nuns \leftarrow nun \cup \{i\epsilon I \mid class(i) = class(nun)\}$
12     $cfs \leftarrow \{genCF(p, n) \mid n \; \epsilon \; nuns\}$
13     $cfs \leftarrow \{cf \mid cf \; \epsilon \; cfs \land validateCF(cf, xc, M)\}$
14     **return** $sort(cfs, key = sim(cf, p))$

15 **def** genCF($p, nun$):
16     $m \leftarrow \{f \mid f \; \epsilon \; matches(p, nun)\}$
17     $d \leftarrow \{f \mid f \; \epsilon \; diffs(p, nun)\}$
18     $cf \leftarrow m \cup d$
19     **return** $cf$ if $cf \neq p$

20 **def** validateCF($cf, xc, M$):
21     **return** $M(cf) = class(xc.x')$

**Algorithm 1:** Generating counterfactuals by reusing the $k$ nearest explanation cases to a target problem, $p$.

---

This new method is a desirable extension of KS20 be-cause it promises better coverage, plausibility and diversity. On *coverage*, generating more counterfactual candidates greatly improves the chances of producing valid counterfactuals and, therefore, should increase the fraction of target problems that can be explained with a good counterfactual. On *plausibility*, this approach has the potential to generate counterfactuals that are even more similar to the target problem than those associated with the nearest explanation case. Finally, on *diversity*, since different explanation cases may rely on different combinations match/difference features, from different explanation cases, the resulting counterfactuals should draw from a more diverse set of difference-features.

## 3.2 Validating Counterfactuals

In a multi-class setting ($n > 2$ classes), there are are least two ways to validate a candidate counterfactual, $cf$. One approach is to look for a *class change* so that $cf$ is considered valid if its predicted class ($M(cf)$) differs from $p$'s class, as mentioned above and used by KS20. Another option, is to confirm that $cf$ has the *same class* as the NUN used to produce it ($M(cf) = class(xc_i.x')$. In this work we use this latter, *same class*, approach (lines 19-20 in Algorithm 1) because it provides for a stronger validation test than the weaker *class change* approach used by KS20; it is harder for a $cf$ to pass the *same class* test because there is only a single valid class, whereas any one of $n - 1$ classes will work in the case of KS20's *class change*. The stronger *same class* test is also more appealing because it constrains the $cf$ to remain within the vicinity of the NUN class used to produce it ($class(xc.x')$), thereby ensuring greater plausibility.

## 4 Evaluation

We evaluate the counterfactuals produced by the $k$-NN approach using 10 well-known ML datasets (see the legends of Figures 1 and 2) with varying numbers of classes, features, instances, in comparison to 2 baselines, and using 3 evaluation metrics (coverage, distance, and diversity). The evaluation was implemented in Python 3.8, on a 72 core Intel Xeon ( 512GB RAM, 1x 1TB SSD, 6x 10TB SATA RAID, 2x NVIDIA 2080 TI), with an total run-time of approximately 60 hours.

## 4.1 Methodology

A form of 10-fold cross-validation is used to evaluate the newly generated counterfactuals, by selecting 10% of the training instances at random to use as test/target problems. Then, the XC case-base is built from a subset of the XCs that are available from the remaining instances; we use at most 2x as many XCs as there are test problems. Finally, any remaining instances, which are not part of any selected XCs, are used to train the underlying classifier; in this case we use a gradient boosted classifier [Friedman, 2002][2], which was found to be capable of generating sufficiently accurate classification performance across the available datasets, but obviously any alternative classifier could be considered.

We use the $k$-NN technique to generate good counterfactuals, varying $k$ and $d$. Since this is a form of endogenous counterfactual generation, we use two variants of the (endogenous) KS20 approach as baselines: (i) a *retrieval only 1NN* variant, which generates a counterfactual from a single XC (and its single NUN), and (ii) a *retrieval & adaptation* variant (*1NN\**) which also uses a single XC, but considers nearby neighbours of the XC's NUN, as a source of extra difference features (KS20's two-step method). This *1NN\** variant is equivalent to our $k$-NN approach with $k = 1$; note, KS20 found the *1NN\** variant to be superior to the *1NN* variant.

Generated counterfactuals are evaluated using 3 different metrics, averaging across the test cases and folds:

- *Test Coverage*: the fraction of test problems that can be associated with at least one good counterfactual, to assess explanatory coverage

- *Relative Distance*: the ratio of the distance (inverse similarity) between the *closest cf* produced and $p$, and the distance between $xc.x$ and $xc.x'$ from the XC used to generate $cf$; relative distance <1 means the $cf$ is *closer* to $p$ than $xc.x$ was to $xc.x'$. This is our proxy measure for reflecting plausibility[3].

- *Feature Diversity*: the fraction of unique difference features in the counterfactuals; thus, a diversity of 0.1 means that 10% of all features appeared as difference features in the counterfactuals generated.

Note, the *1NN* baseline always has the same diversity as *1NN\**, as a single XC is reused (one set of differences) and so we do not separately report diversity for *1NN*.

## 4.2 Results

Figures 1 and 2 show the results for $1 \leq k \leq 100$ with $d = 2$ and $d = 3$. Performance on each dataset is represented as a separate line graph with statistical significance encoded as follows. If the difference between two successive points (for a given dataset) is statistically valid ($p < 0.05$) then the points are connected by a solid line, otherwise they are connected by a dashed line; for coverage we use a z-test of proportions and for relative distance and diversity we use a *t-test*. Separately, if a marker is filled then it means that the difference between its value and the *1NN\** baseline is statistically significant (also for $p < 0.05$). Notice that the x-axis is not strictly linear, to provide a greater level of detail for $k \leq 10$.

In Figure 1(a) we can see how the ability to produce good counterfactuals increases with $k$ up to a point, which depends on the number of available XCs for each dataset. In all datasets, coverage for $k > 1$ is significantly greater than the KS20-baseline, and coverage increases to more than 80% of target problems in all cases for a large enough $k$. On average, the current $k$-NN approach is able to increase coverage by almost a factor of 2, compared to the KS20 *1NN\** baseline, as indicated by the *relative improvement* values for coverage in Figure 1(d); the approximate values for $k$ shown, indicate when this maximum coverage is achieved.

---

[2]We used the SciKitLearn implementation with a deviance loss function, a learning rate of 0.1, and 100 boosting stages.

[3]As this is an endogenous technique the out-of-distribution metrics sometimes used in evaluating exogenous techniques are not germane.

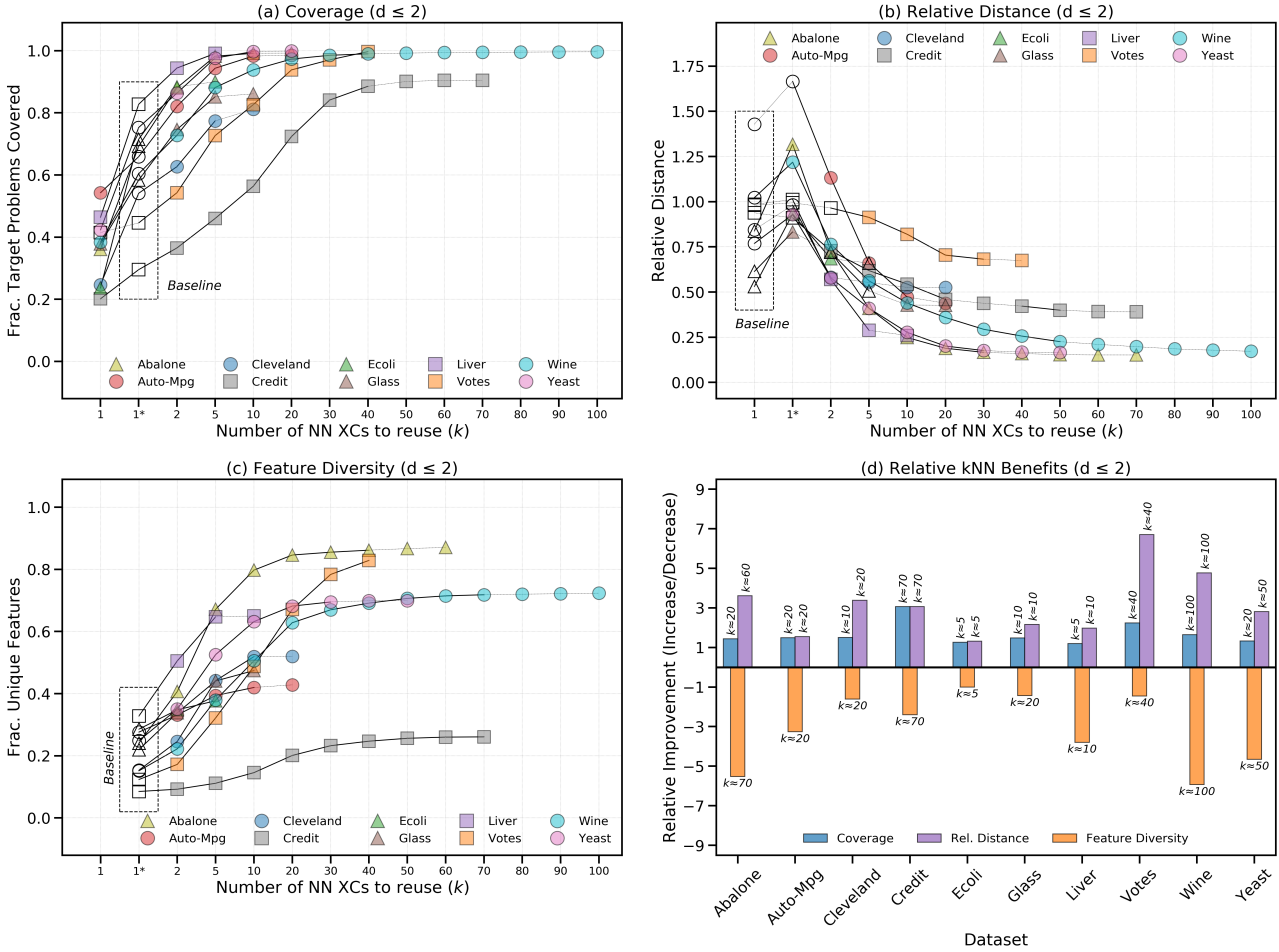

Figure 1: Counterfactual evaluation results for $d \leq 2$: (a) counterfactual coverage, (b) mean relative distance, and (c) counterfactual diversity along with the relative improvements (d) compared to the *1NN/1NN\** baseline as appropriate.

In Figure 1(b) we see these coverage improvements also offer statistically significant reductions in relative distance, for increasing $k$, this time compared with *1NN*, since it offers better relative distance than *1NN\**. Thus, by using additional explanation cases, even those that are further away from the test/target problem, we can generate valid counterfactuals that are even closer to the target. The increase in relative distance for *1NN\**, compared with the *1NN*, is due to the significant increase in coverage offered by *1NN\**, which means there are far more valid counterfactuals participating in the relative distance calculations. Once again, in Figure 1(d) we show a relative improvement (decrease) in these distances (compared with *1NN* baseline): on average there is a 3x decrease in relative distance. This is usually achieved for a larger value of $k$ than the best coverage, which highlights the benefits of continuing the search beyond an initial valid counterfactual.

Finally, the diversity results are presented in Figure 1(c), showing significant improvements in feature diversity with increasing $k$, although not every dataset produces counterfactuals with high levels of diversity. For example, in Figure 1(c) the counterfactuals produced for *Credit* only include 25% of the available features, so most features don't serve as difference features. On the other hand, the counterfactuals produced for *Abalone* include over 80% of features as their difference features, while datasets such as *Auto-MPG, Cleveland, and Glass* achieve more moderate levels of diversity with 45% to 50% feature participation. Nevertheless, these are considerable improvements ( 3x) compared to the diversity of the baseline (*1NN\**) approach, as per Figure 1(d).

The $d = 3$ results in Figure 2, though not discussed in detail, show similar trends: coverage and diversity in-creases with $k$ while relative distance decreases, and in all cases the best results for $k$-NN demonstrate significant improvements over the KS20-baselines.

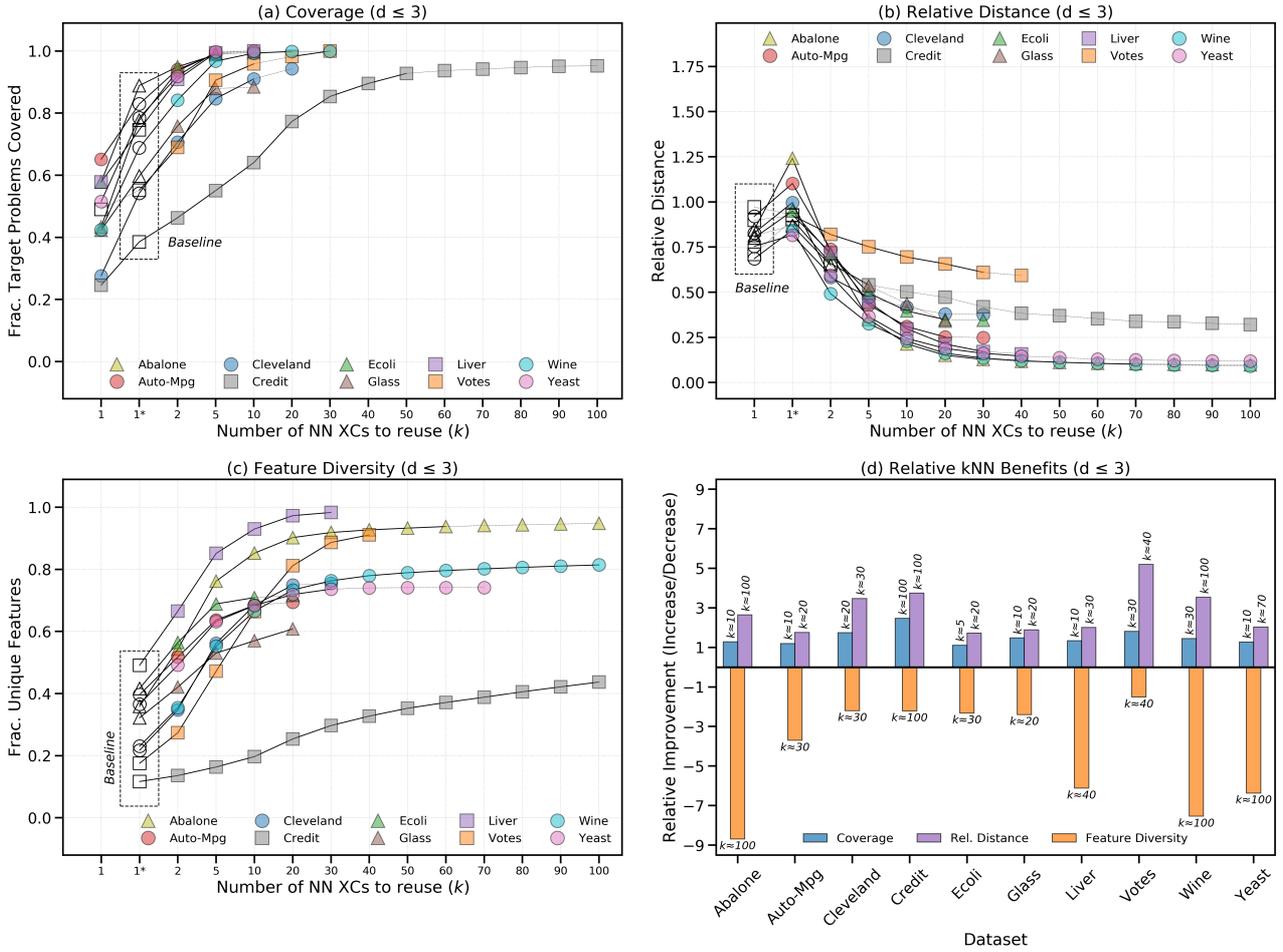

Figure 2: Counterfactual evaluation results for $d \leq 3$: (a) counterfactual coverage, (b) mean relative distance, and (c) counterfactual diversity along with the relative improvements (d) compared to the *1NN/1NN\** baseline as appropriate.

## 5 Conclusions

Counterfactuals are playing an increasingly important role in Explainable AI research because they can be more causally informative than alternative factual forms of explanation. However, useful native counterfactuals – those that are similar to a target problem but differ in only a few features – can be rare in many real-world settings, leading some researchers to propose *exogenous* techniques for generating synthetic counterfactuals [Wachter *et al.*, 2017; Dandl *et al.*, 2020; Mothilal *et al.*, 2020]. While such exogenous techniques have shown that synthetic counterfactuals can be produced, they often rely on features that cannot be guaranteed to occur naturally, which may limit their explanatory-utility for end-users. In response, other researchers have advanced alternative *endogenous* techniques for generating counterfactuals, based on features that naturally occur among the instances of a dataset.

The main contribution of this work is a novel approach to counterfactual generation that significantly advances previous proposals on endogenous counterfactual generation, arguably in a more elegant manner. The second main contribution lies in its systematic and extensive comparative testing of endogenous techniques to demonstrate the optimal parameters for current and previous methods, across a wide range of benchmark datasets.

As with any research, there are limitations that invite future directions. We have focused on classification tasks, but in principle the approach should be equally applicable to prediction tasks. The current evaluation focuses on a *like-for-like* comparison with endogenous counterfactual generation methods. Further comparisons are planned, to compare exogenous and endogenous techniques more directly. Perhaps more importantly, as is the case with other counterfactual techniques, though we have provided an offline analysis of counterfactual quality, we have not yet evaluated the counterfactuals produced *in situ*, as part of a real live-user explanation setting. This will be an important part of future research, as the utility of any counterfactual generation technique will depend critically on the nature of the counterfactuals produced and their informativeness as explanations to "real" end-users. But, to be positive, the current tests identify the optimal versions of these methods that need to be used in such future user studies.